\documentclass[runningheads]{llncs}

\usepackage{graphicx}
\usepackage{comment}
\usepackage{amsmath,amssymb}
\usepackage{color}
\usepackage{url}
\usepackage{hyperref}
\usepackage{subcaption}

\newcommand{\Variable}[1]{\mathrm{\MakeLowercase{{\textit{#1}}}}}
\newcommand{\Variablei}[2]{{{\Variable{#1}_{#2}}}}
\newcommand{\Variableij}[3]{{{\Variable{#1}_{#2}^{#3}}}}
\newcommand{\RandomVariable}[1]{\mathrm{\MakeUppercase{{\textit{#1}}}}}
\newcommand{\RandomVariablei}[2]{{{\RandomVariable{#1}_{#2}}}}
\newcommand{\RandomVariableij}[3]{{{\RandomVariable{#1}_{#2}^{#3}}}}
\newcommand{\Set}[1]{{\mathrm{\MakeUppercase{\mathcal{{#1}}}}}}
\newcommand{\Seti}[2]{{{\Set{#1}_{#2}}}}

\newcommand{\Prob}[1]{{{\Pr[#1]}}}

\newcommand{\ProbGiven}[2]{{{\Pr[#1|#2]}}}
\newcommand{\Function}[1]{{\mathrm{#1}}}

\newcommand{\Rot}[0]{{\mathrm{R}}}
\newcommand{\Trans}[0]{{\vec{\mathrm{t}}}}
\newcommand{\Map}[0]{{\mathrm{map}}}
\newcommand{\Query}[0]{{\mathrm{q}}}

\newcommand{\vi}[0]{{\Variable{i}}}
\newcommand{\vj}[0]{{\Variable{j}}}
\newcommand{\vk}[0]{{\Variable{k}}}
\newcommand{\vl}[0]{{\Variable{l}}}
\newcommand{\vm}[0]{{\Variable{m}}}
\newcommand{\vn}[0]{{\Variable{n}}}

\newcommand{\vr}[0]{{\Variable{r}}}

\newcommand{\vv}[0]{{\Variable{v}}}

\newcommand{\vx}[0]{{\Variable{x}}}
\newcommand{\vy}[0]{{\Variable{y}}}

\newcommand{\vnr}[0]{{\Variablei{n}{\vr}}}
\newcommand{\vfr}[0]{{\Variablei{f}{r}}}

\newcommand{\vvp}[0]{{\Variablei{v}{p}}}
\newcommand{\vvi}[0]{{\Variablei{v}{i}}}
\newcommand{\vfq}[0]{{\Variablei{f}{q}}}
\newcommand{\pinvotev}[1]{{\Variableij{p}{\mathrm{in\_vote}}{#1}}}
\newcommand{\poutvote}[0]{{\Variablei{p}{\mathrm{out\_vote}}}}

\newcommand{\sI}[0]{{\Set{I}}}
\newcommand{\sO}[0]{{\Set{O}}}
\newcommand{\sM}[0]{{\Set{M}}}
\newcommand{\sR}[0]{{\Set{R}}}

\newcommand{\sV}[0]{{\Set{V}}}

\newcommand{\sFq}[0]{{\Seti{F}{q}}}

\newcommand{\sFR}[0]{{\Seti{F}{\sR}}}
\newcommand{\sVinlier}[0]{{\Seti{V}{\sI}}}

\newcommand{\fB}[0]{{\Function{B}}}

\newcommand{\expected}[1]{{\mathbb{E}(#1)}}

\newcommand{\rI}[0]{{\RandomVariablei{N}{\sI}}}
\newcommand{\rIi}[1]{{\RandomVariableij{N}{\sI}{#1}}}
\newcommand{\rO}[0]{{\RandomVariablei{N}{\sO}}}
\newcommand{\rOi}[1]{{\RandomVariableij{N}{\sO}{#1}}}
\newcommand{\rM}[0]{{\RandomVariablei{N}{\sM}}}
\newcommand{\rMi}[1]{{\RandomVariableij{N}{\sM}{#1}}}

\begin{document}

%%%%%%%%%%%%%%%%%%%%% Add submission id, track, and title. %%%%%%%%%%%%%%%%%%%%%

% Replace with your title
\title{Model-Based Parameter Optimization for Ground Texture Based Localization Methods}


%	\author{First Author\inst{1}\orcidID{0000-1111-2222-3333} \and
%	Second Author\inst{2,3}\orcidID{1111-2222-3333-4444} \and
%	Third Author\inst{3}\orcidID{2222--3333-4444-5555}}
%	
%	\authorrunning{F. Author et al.}
%	% First names are abbreviated in the running head.
%	% If there are more than two authors, 'et al.' is used.
%	
%	\institute{Princeton University, Princeton NJ 08544, USA \and Springer Heidelberg, Tiergartenstr. 17, 69121 Heidelberg, Germany
%	\email{lncs@springer.com}\\
%	\url{http://www.springer.com/gp/computer-science/lncs} \and ABC Institute, Rupert-Karls-University Heidelberg, Heidelberg, Germany\\
%	\email{\{abc,lncs\}@uni-heidelberg.de}}

\titlerunning{Model-Based Parameter Optimization for GTBL Methods}
\author{Jan Fabian Schmid\inst{1,2} \and
	Stephan F. Simon\inst{1} \and
	Rudolf Mester\inst{3,2}}
\authorrunning{J.F. Schmid et al.}	
\institute{Robert Bosch GmbH, Hildesheim, Germany\\
	\email{SchmidJanFabian@gmail.com} \and
	VSI Lab, CS Dept., Goethe University, Frankfurt am Main, Germany \and
	Norwegian Open AI Lab, CS Dept., NTNU Trondheim, Norway
}

\maketitle              % typeset the header of the contribution

%%%%%%%%%%%%%%%%%%%%%%%%%%%%%%%%%%%%%%%%%%%%%%%%%%%%%%%%%%%%%%%%%%%%%%%%%%%%%%%%

% !TeX spellcheck = en_US
\begin{abstract}
	A promising approach to accurate positioning of robots is ground texture based localization.
	It is based on the observation that visual features of ground images enable fingerprint-like place recognition.
	We tackle the issue of efficient parametrization of such methods,
	deriving a prediction model for localization performance,
	which requires only a small collection of sample images of an application area.
	In a first step,
	we examine whether the model can predict the effects of changing one of the most important parameters of feature-based localization methods:
	the number of extracted features.
	We examine two localization methods,
	and in both cases our evaluation shows that the predictions are sufficiently accurate.
	Since this model can be used to find suitable values for any parameter,
	we then present a holistic parameter optimization framework,
	which finds suitable texture-specific parameter configurations,
	using only the model to evaluate the considered parameter configurations.
\end{abstract}

\section{Introduction}
Accurate localization capabilities are a necessary precondition for many tasks of autonomous vehicles and moving robots.
A low-cost approach to this is ground texture based localization,
which uses only a single downward-facing camera.
This approach has some %evident
advantages over approaches using forward-facing cameras: %or surround-view cameras:
it does not have any issues with occlusions of the surrounding,
it works in environments without static landmarks,
and it can be made independent of the surrounding lighting.
Previous work has shown that reliable high-accuracy pose estimates can be achieved with ground texture based 
localization on typical ground texture types, such as asphalt and concrete~\cite{chenstreetmap},\,\cite{Fang_intelligent-vehicles2},\,\cite{Kelly_AGV},\,\cite{Kozak_Ranger},\,\cite{Nagai_Path_Tracking},\,\cite{Schmid_GTBL},\,\cite{Zhang_High-Prec-Localization}.

Realizing optimal performance of localization methods typically requires the choice of a variety of parameters,
such as the number of considered visual features per image.
Finding optimal parameter settings is often a time-consuming process,
in which many possible choices are considered.
It is therefore desirable to predict the localization performance without having to extensively evaluate the method.

We consider a scenario where an agent (e.g. a mobile robot) creates a map of the application area.
Subsequently,
the agent should be able to localize itself anywhere in this environment even without prior knowledge about its whereabouts.
Therefore, we require the examined methods to perform \emph{absolute} localization,
i.e. they estimate the camera pose in the map coordinate system,
and we require that the method is able to perform \emph{global} localization,
i.e. it does not require a prior localization estimate,
but considers all possible poses simultaneously.
Having such a self-contained method means that localization can be performed entirely with the images of a single 
downward-facing camera.

We introduce a prediction model for the success rate of two state-of-the-art ground texture based localization methods~\cite{Zhang_High-Prec-Localization},\,\cite{Schmid_GTBL}.
Our model requires only a few test images of the application area,
for each of which we test if the localization methods succeed in estimating its pose in respect to the others.
This allows to quickly determine the \emph{local} localization performance.
Assuming similar properties over the application area,
the model then uses the local knowledge to estimate the expected \emph{global} localization performance.
Based on the predicted global performance,
the agent is then able to optimize the localization %method according to the challenges of the current ground texture type.
for the current ground texture type.
In comparison,
a simple black-box approach to parameter evaluation would directly evaluate the global localization performance,
which is more accurate but has significantly higher computational cost,
because many more images are processed.
Therefore, besides a deeper understanding of the factors that lead to successful localization of the examined localization methods,
our model-based performance evaluation allows to consider more parameter configurations in the same amount of time.
Accordingly, the prediction model enables faster deployment of agents in new application areas.

This paper contributes the first model-based performance evaluation approach for feature-based 
localization methods using images of a downward-facing camera.
We evaluate its predictive power,
and examine its suitability to be used in a parameter optimization framework.

% TODO structure of paper?
%The structure of the paper is as follows.
%First, in Section~\ref{sec:problem_statement},
%we provide a formal definition of the properties of the examined localization methods.
%Section~\ref{sec:model} derives the prediction model and explains how to apply it for parameter optimization.
%Subsequently, Section~\ref{sec:evaluation} introduces the experiments to evaluate the predictive power of the model.
%The results of these experiments are evaluated in Section~\ref{sec:results}.
%Finally, Section~\ref{sec:conclusion} concludes with a short discussion of our results.

\section{Related Work}
State-of-the-art methods solving the introduced localization task rely on the identification of visual feature correspondences to 
determine the pose of a query image relative to the mapped reference 
images~\cite{chenstreetmap},\,\cite{Schmid_GTBL},\,\cite{Zhang_High-Prec-Localization}.
They extract \emph{features}, i.e. characterizations of image patches. %, from the images.
Features consist of two parts, (1) a specification of the place of the image patch,
which we refer to as \emph{keypoint object},
that is composed of a keypoint,
defining the image coordinates of the corresponding region, its orientation, and size;
and (2) a feature descriptor that summarizes the image content of the patch,
e.g. as a binary or real-valued vector.
The considered feature-based localization methods operate as follows.
\begin{enumerate}
	\item \emph{Mapping phase:} poses of the reference images
	are determined for a joint map coordinate system,
	e.g. using an image stitching process. %, which aligns the images to each other.
	Assuming a pinhole camera model with constant height, perpendicular to the ground,
	image poses can be described as standard Euclidean transformations of $\vx$ and $\vy$-coordinates and an orientation angle.
	Subsequently, features of reference images are extracted in two steps.
	(a) A keypoint detector determines keypoint objects with the goal of finding similar image patches in the mutual regions of overlapping images.
	(b) Feature descriptors are computed for those keypoint objects.
	For a given distance metric,
	descriptors should be close for corresponding keypoint objects and distant otherwise.
	\item \emph{Localization phase:} after mapping,
	the method estimates the pose of a query image. % in the map coordinate system.
	Features are extracted from the query image similarly as for the reference images.
	Then, in a feature matching step, correspondences between query and reference features are proposed,
	and finally, based on the correspondence candidates, the query image pose is estimated.
\end{enumerate}

Feature-based methods were shown to achieve high localization success rates on ground images~\cite{Schmid_GTBL},
where localization attempts are considered successful,
if the estimated query image pose is close to the ground truth pose.
We adopt the thresholds as in~\cite{Zhang_High-Prec-Localization} and \cite{Schmid_GTBL} to define closeness:
the position error threshold is $30$\,pixels and the orientation error threshold is $o_t=1.5$\,degrees.

\subsubsection{Approaches to Parameter Optimization}
Apart from the success rate, computational effort and memory consumption of the localization method should be optimized.
Optimizing parameter configurations for these goals is a complex task that requires to make trade-off decisions.
So far, ground texture based localization methods have been parametrized based on extensive empirical 
evaluation of possible parameter values~\cite{Schmid_Survey},\,\cite{Schmid_GTBL},
treating the method as a black-box.

An alternative was developed by Mount et al.~\cite{Mount_CoverageSelection}.
They proposed a method to automatically determine a suitable trade-off between camera coverage area and localization performance,
using only a few %pairs of
aligned test images.
Similarly, our approach requires only a few test images of the application area,
but it allows to optimize any parameter of feature-based localization methods.
Since the camera coverage is not a relevant parameter for the state-of-the-art
localization methods being considered in this work,
we cannot compare with Mount et al.'s approach.
%Also, the approach of Mount et al.~\cite{Mount_CoverageSelection} is not model-based, 
%it involves evaluating all considered parameter values on the test images,
%while our approach can avoid that for some parameters,
%like the number of extracted features per image.

\section{Localization Method Properties}
\label{sec:problem_statement}
We define the properties of the examined ground texture based methods for global localization.
The available input consists of a set of reference images $\sR$ that have already been properly aligned with each other,
%e.g. using an image stitching procedure,
and a query image $\Query$ that we want to localize,
which was also recorded in the mapped area.

We consider feature-based methods.
The method extracts a set of reference features $\sFR$ from the reference images ($\vnr$ per image),
and a set of query features $\sFq$ from the query image.
Extracted keypoint objects specify the orientation of their image patches.
Also, keypoint objects of query features specify their position in the query image,
while keypoint objects of reference features specify their position in the map.
Then, a matching method proposes a set of matches $\sM$ as possible correspondences between the feature sets.
Every match $\vm\in\sM$ is a pair consisting of a query feature and a reference 
feature $\vm = (\vfq\in\sFq, \vfr\in \sFR)$.
Finally, a pose estimation method uses the matches to estimate the Euclidean transformation 
$[\Rot|\Trans]_\Query^\Map$
projecting the query image onto the map. %, e.g. using a RANSAC procedure.

\begin{figure}[tb]
	\tiny 
	\centering
	\includegraphics[width=0.495\columnwidth]{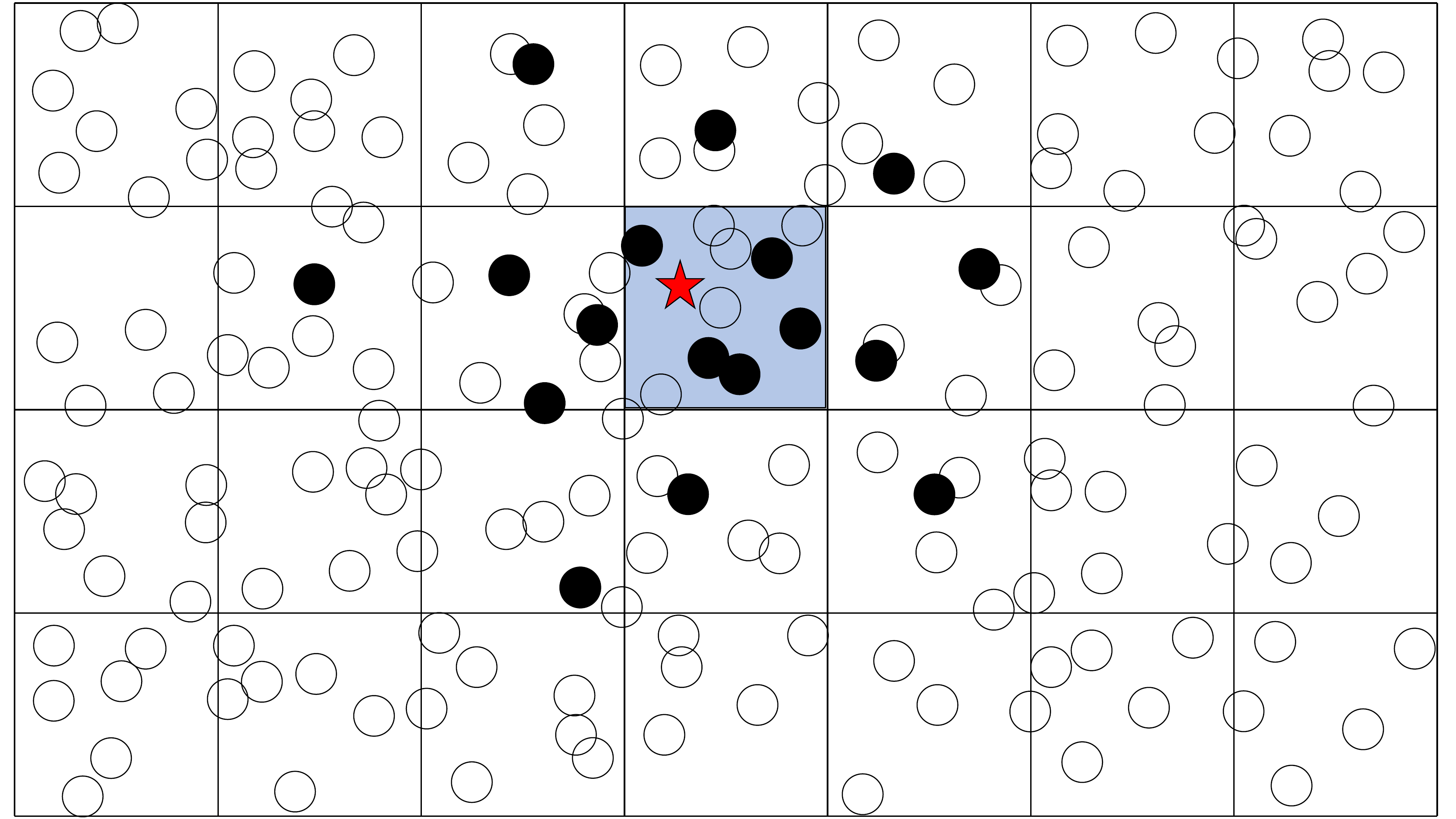}
	\includegraphics[width=0.495\columnwidth]{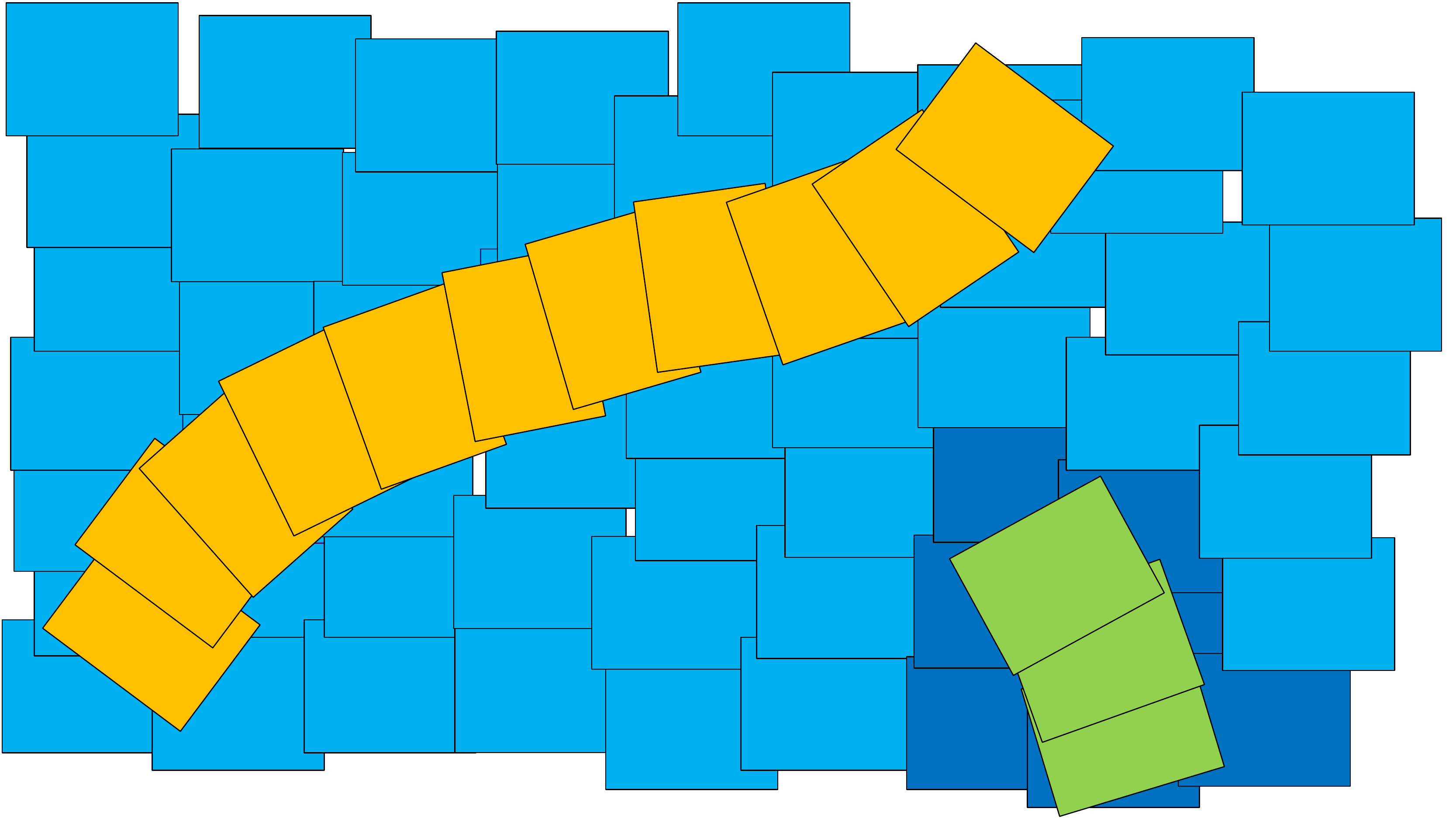}	
	\caption{
		\textit{Left:} Illustration of the voting procedure.
		A 2D histogram splits the map into equally sized voting cells.
		Every feature match votes for the query image position (circles).
		Some matches, the inliers, represent correct feature correspondences (solid circles)
		that enable correct query image pose estimation.
		Inlier votes concentrate close to the true query image position (red star),
		while outlier votes (transparent circles) are distributed randomly.
		Only the matches voting for the cell that received most votes (blue),
		which we call \emph{voting peak},
		are used for the subsequent pose estimation step.
		\textit{Right:} Illustration of the available images.
		A large set of partially overlapping reference images (blue) is used for mapping.
		Independently recorded query images (orange) are to be localized globally.
		For parametrization, we only evaluate local performance,
		using a few additional test images (green) and
		their overlapping reference images (dark blue).
	}
	\label{fig:database_and_voting_map}	
\end{figure}

In addition to the previously described pipeline,
we assume that,
\emph{prior} to the pose estimation step,
examined methods employ a \emph{voting procedure} for spatial verification of matches
(illustrated in Fig.~\ref{fig:database_and_voting_map} (left)).
This allows to reject incorrect matches (outliers),
which is useful as feature-based global localization methods on ground texture were shown to suffer from large 
quantities of outliers~\cite{Zhang_High-Prec-Localization}, \cite{Schmid_GTBL}.
One approach to outlier rejection is spatial verification of feature matches, using Hough voting 
approaches~\cite{Avrithis_SpatialVerification},\,\cite{Schonberger_SpatialVerification},\,\cite{Zeisl_SpatialVerification}.
The idea to use a voting procedure for spatial verification of feature matches from ground textures was proposed by 
Zhang et al.~\cite{Zhang_High-Prec-Localization}.
Their proposed technique exploits the fact that every match of ground features represents a query image pose estimation.
Given a match $\vm = (\vfq\in\sFq, \vfr\in \sFR)$,
we can derive the transformation $[\Rot|\Trans]_\Query^\vfq$ that maps from the query image pose to the pose of $\vfq$,
the transformation $[\Rot|\Trans]_\vfr^\Map$ that maps from the reference feature pose to the map coordinate system,
and the transformation $[\Rot|\Trans]_\vfq^\vfr$ that maps from the query feature pose to the reference 
feature pose,
which, due to the assumed (correct) correspondence of $\vfq$ and $\vfr$,
is estimated to be the identity.
This allows to compute the query image pose as
\begin{equation}
[\Rot|\Trans]_\Query^\Map = [\Rot|\Trans]_\vfr^\Map[\Rot|\Trans]_\vfq^\vfr[\Rot|\Trans]_\Query^\vfq = 
[\Rot|\Trans]_\vfr^\Map[\Rot|\Trans]_\Query^\vfq\,.
\end{equation}

There is a subset of matches that can be considered correct,
the set of \emph{inlier} matches $\sI$,
while the remaining incorrect matches $\sO$ are \emph{outliers} ($\sM=\sI\cup\sO$).
We consider a match to be correct,
if it can be used to determine the correct query image pose (details in the supplementary material).
In order to reject outliers,
the voting procedure proceeds as follows.
Every proposed match votes for the position of the query image in the map coordinate system,
using the translation component of their corresponding pose estimate,
while ignoring the potentially inaccurate orientation estimate.
Votes cast into a similar area are summarized using a 2D histogram,
i.e. a grid of equally sized voting cells.
We call the voting cell with most votes the \emph{voting peak}.
Only the matches voting for the voting peak will be considered during the subsequent pose estimation step,
as we expect some of them to be inliers.
Even though inliers, in particular due to inaccuracies in the keypoint object orientations,
are not necessarily voting for positions precisely at the true query image position,
we expect them to concentrate close to it,
while outlier votes are scattered randomly over the voting histogram.

\section{Examined Localization Methods}
We will evaluate the predictive power of the prediction model for the localization methods:
Micro-GPS~\cite{Zhang_High-Prec-Localization} and an adaptation of it developed by Schmid et al.~\cite{Schmid_GTBL}.
They have the properties described in Section~\ref{sec:problem_statement} and
were shown to be the state of the art for ground texture based global localization~\cite{Schmid_GTBL},
being able to reach almost perfect success rate on asphalt, carpet, and tiles
for application areas consisting of $2000$ to $4000$ images from the
database of Zhang et al.~\cite{Zhang_High-Prec-Localization}.

\subsection{Micro-GPS of Zhang et al.}
Micro-GPS\footnote{\url{https://microgps.cs.princeton.edu/}}
was developed by Zhang et al.~\cite{Zhang_High-Prec-Localization}.
%Its code is publicly available \footnote{\url{https://microgps.cs.princeton.edu/}}.
%
It extracts SIFT~\cite{Lowe_SIFT2} features from every reference image.
%using SiftGPU\footnote{\url{https://github.com/pitzer/SiftGPU}}.
Only a subset of $\vnr$ randomly sampled SIFT features are kept per reference image,
with $\vnr=50$ in the original implementation.
Then, principal component analysis (PCA) is employed to reduce the feature descriptor dimensionality to $8$ or $16$.
We employ 16-dimensional descriptors,
as the authors observed better performance with it~\cite{Zhang_High-Prec-Localization}.
Subsequently, with help of the FLANN library~\cite{Muja_FLANN},
all reference features are stored for efficient feature matching in an approximate nearest neighbor (ANN) search 
structure.
In the localization phase,
SIFT features are extracted from the query image and their descriptor values are again mapped to $16$ dimensional 
real-vectors using PCA,
but all detected features are used in this case.
Every query image feature is matched with its ANN among the reference features,
using the previously created ANN search structure.
The voting procedure is applied for spatial verification of these matches,
using a histogram cell size of $50 \times 50$ pixels ($8 \times 8$\,mm).
Finally, based on the matches of the voting peak,
the query image pose is estimated in a RANSAC procedure.

\subsection{Localization Method of Schmid et al.}
Schmid et al. introduced an adaptation of Micro-GPS~\cite{Schmid_GTBL},
improving on some of its shortcomings.
This method substitutes the ANN search structure of Micro-GPS with a novel feature matching strategy called 
\emph{identity matching},
which matches only features with identical descriptor values.
In order to apply this feature matching strategy,
they use LATCH~\cite{Levi_LATCH} to compute binary descriptor vectors for the extracted SIFT keypoint objects,
but keep only the first $15$ bit of the descriptors.
The number of descriptor dimensions kept is a trade-off between the number of generated outlier matches,
that increases when considering less bits,
and the number of missed inliers,
that increases when considering more bits.
The OpenCV 4.0 library~\cite{opencv_library} is employed to detect $\vnr=850$ SIFT keypoint objects per image,
and to describe them with the $15$-bit LATCH descriptors.
In comparison to an ANN search structure for feature matching,
identity matching has some crucial advantages:
the ANN search structure is recomputed whenever the set of reference features $\sFR$ changes,
e.g. when a reference image is added, removed, or updated,
which is not necessary when using identity matching.
This is because it performs matching efficiently on an image-to-image basis.
For the same reason,
the method is able to restrict the set of considered reference images during the feature 
matching step to those images that are close to a query image pose estimate if available.
This significantly reduces computational effort and
improves localization success rates on more difficult ground textures~\cite{Schmid_GTBL}.

Further, the method uses the voting procedure with a histogram cell size of $75 \times 75$ pixels ($12 \times 
12$\,mm) for spatial verification of the obtained matches,
and then estimates the query image pose using RANSAC,
similar to Micro-GPS.

\section{The Prediction Model}
\label{sec:model}
A complete derivation of our prediction model for the success rate of a method with the properties described in Section~\ref{sec:problem_statement} is given in the supplementary material.

The main assumption of the model is that localization succeeds if among the matches voting for the voting peak $\vvp$ are at least two inliers,
denoted as $\rIi{\vvp}\ge2$.
According to our results,
this is an accurate assumption.
For both evaluated localization methods,
the success rate is greater than $99.3\%$ for localization attempts that hold the condition,
while the success rate over all localization attempts is $89.9\%$ for Micro-GPS,
and $94.3\%$ for Schmid et al.'s method.
Also, not a single localization attempt that does not hold the condition succeeded.

\begin{figure}[t]
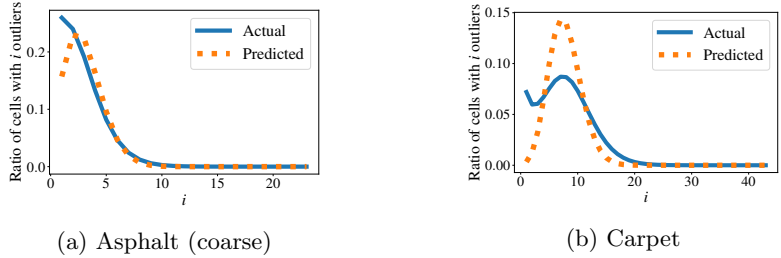

	\vspace*{0.125cm}
	\begin{subfigure}[c]{0.346\columnwidth}
		\includegraphics[width=1\columnwidth]{src/GCPR2020_IdentityMatchingApproach_nof_correct_matches_gt_cell_one_of_three/%
			hist_nof_outliers_per_voting_cell_prob_IdentityMatchingApproach_coarse_10.pdf}
		\subcaption{Asphalt (coarse)}
	\end{subfigure}\hfil
	\begin{subfigure}[c]{0.346\columnwidth}
		\includegraphics[width=1\columnwidth]{src/GCPR2020_IdentityMatchingApproach_nof_correct_matches_gt_cell_one_of_three/%
			hist_nof_outliers_per_voting_cell_prob_IdentityMatchingApproach_carpet_10.pdf}
		\subcaption{Carpet}
	\end{subfigure}
	\caption{Actual outlier vote histograms observed over $500$ global localization attempts and its predicted binomial distribution.
		For two ground texture types,
		the graphs show the ratio of voting cells that received certain amounts of outliers.}
	\label{fig:schmid_dist_outliers}	
\end{figure}

A second important assumption is made about the spatial distribution of outlier votes on the voting map.
Here, we assume to have complete spatial randomness (CSR),
i.e. the probability $\poutvote$ of any outlier match $\vm\in\sO$,
casting a vote on the voting cell $\vv$ is the same for any voting cell $\vv\in\sV$.
% TODO Binomial distribution
For Schmid et al.'s method,
we compare the actual outlier distributions with the ones predicted based on the CSR assumption.
Fig.~\ref{fig:schmid_dist_outliers} presents the results for two representative textures.
We find our predicted outlier distribution to be sufficiently accurate.
While we systematically underestimate the number of voting cells that receive only very few outlier votes,
the predicted numbers of voting cells with larger amounts of outlier votes is more accurate.
In practice, only the voting cells that received most votes influence the localization success rate.
Therefore, the CSR assumption seems to be sufficiently accurate to predict success rates.

Our final success rate prediction model is the following:
\begin{equation}
\label{eq:model}
\Prob{\rIi{\vvp}\ge2} = 1 - \left(\prod_{\vvi\in\sVinlier}\left[1 - \left(\Prob{(\rIi{\vvi}\ge2)\cap(\rMi{\vvi}=\rMi{\vvp})}\right)\right]\right).
\end{equation}
It means that the predicted success rate corresponds to the probably that,
among the voting cells that received inlier votes $\sVinlier$,
there is at least one voting cell $\vvi\in\sVinlier$,
that obtained two or more inlier votes ($\rIi{\vvi} \ge 2$) while it also obtained the most votes of any voting cell, i.e. it obtained the number of votes that the voting peak $\vvp$ obtained: $\rMi{\vvi}=\rMi{\vvp}$.

\subsection{Application of the Prediction Model}
\label{sec:model:application}
To use the model,
we require some empirical observations:
$|\sV|$, the number of voting cells with at least one vote;
$|\sFq|$, the number of extracted query image features;
$\rM$, the number of matches;
$\rO$, the number of outlier matches;
$\sVinlier$, the set of voting cells that received inlier votes;
for all $\vvi\in\sVinlier$, $\pinvotev{\vv}$, the probability that a query image feature generates an inlier vote on $\vvi$.

These model parameter values may vary on every localization attempt.
Therefore, we determine expected values $\expected{\cdot}$ of these parameters for the prediction.

$|\sV|$ scales with the coverage area size,
and is not strongly dependent on the texture type.
To estimate $\expected{|\sV|}$,
we take the average value of $|\sV|$ per reference image from evaluations on other application areas,
and multiply it with the number of reference images of the current application area.

To find appropriate values for the remaining model parameters,
we make use of the small collection of test images.
The test images consist of a series of consecutively recorded query images,
and some additional overlapping images.
For every query image among the test images,
we perform two evaluations in form of localization attempts using the respective localization method with the examined parameter configuration:
the first attempt is part of the \emph{inlier evaluation},
where we use all available overlapping images as reference images;
and in a second attempt, for the \emph{outlier evaluation}, the same query image is used,
but randomly selected non-overlapping images are used as reference images.

We estimate $\expected{|\sFq|}$ as the average number of extracted query image features from both inlier and 
outlier evaluations.

We estimate $\expected{\rO}$ as the average number of proposed matches during outlier evaluation,
because all matches here are outliers.
However,
for Schmid et al.'s method,
where the number of matches scales with the application area size,
we measure the average number of outliers per voting cell with at least one vote,
and then estimate $\expected{\rO}$ through multiplication of that value with $\expected{|\sV|}$.

The inlier evaluation allows us to estimate the number of inliers.
We count how many inliers we find on average on the voting cell with most inliers $\overline{\vn_1}$,
the voting cell with second most inliers $\overline{\vn_2}$,
and so on.
This is done for all voting cells that received at least one inlier vote.
Accordingly, we approximate that there will be a voting cell $\vv_1\in\sVinlier$ with 
$\expected{\rIi{\vv_1}}=\overline{\vn_1}$
and $\expected{\pinvotev{\vv_1}}=|\sFq|/\overline{\vn_1}$,
a voting cell $\vv_2\in\sVinlier$ with $\expected{\rIi{\vv_2}}=\overline{\vn_2}$ and 
$\expected{\pinvotev{\vv_2}}=|\sFq|/\overline{\vn_2}$,
and so on.
Furthermore, we estimate $\expected{\rI}=\sum_{\vvi\in\sVinlier}\expected{\rIi{\vvi}}$,
and therefore $\expected{\rM} = \expected{\rO}+\expected{\rI}$.

\section{Evaluation}
\label{sec:evaluation}
In Section~\ref{sec:evaluation:features},
we will first evaluate the suitability of our model to predict the success rate depending on the number of extracted features.
Subsequently, in Section~\ref{sec:evaluation:optimization_framework},
we introduce and evaluate a parameter optimization framework which uses the model to evaluate parameter configuration candidates.

For both evaluations,
we use the image database of Zhang et al.~\cite{Zhang_High-Prec-Localization}.
It contains six ground textures types, i.e. six application areas,
recorded by a mobile robot equipped with a Point Grey camera (example images are shown in the supplementary material).
To determine the actual global localization success rates,
for each texture,
we use their corresponding $2000$ to $4000$ partially overlapping recordings as reference images
(blue in Fig.~\ref{fig:database_and_voting_map} (right)),
and use $500$ separate recordings as query images (orange in Fig.~\ref{fig:database_and_voting_map} (right)).

In order to predict the value of the global success rate for individual application areas,
as described in Section~\ref{sec:model:application},
we use a set of test images.
For this, we evaluate localization attempts on $10$ additional sequentially recorded query images
(green in Fig.~\ref{fig:database_and_voting_map} (right)),
each with a local map of $10$ reference images
(dark blue in Fig.~\ref{fig:database_and_voting_map} (right)).
For inlier evaluation, we select the $10$ closest reference images of the respective query images,
and, for outlier evaluation,
$10$ randomly selected reference images without overlap with the query image.

\subsection{Predicting the Success Rates for Varying Numbers of Features}
\label{sec:evaluation:features}
Generally, using the procedure described in Section~\ref{sec:model:application},
the model can be used to find suitable values for any parameter.
However, two of the most important parameters,
directly influencing the computational effort and memory consumption,
are the number of extracted features per reference image $\vnr$ and per query image $|\sFq|$.
As $|\sFq|$ is not a free parameter of Micro-GPS,
we will focus on $\vnr$.

To find suitable values for $\vnr$,
we exploit an advantage of our model-based parameter evaluation strategy:
if we are able to estimate the impact of $\vnr$ on the model input values,
we can predict the success rate for varying $\vnr$ values without even having to evaluate them on the test images.
Therefore, we evaluate the localization methods on the test images using only a single value of $\vnr$,
namely the value suggested by the corresponding authors.
Then, we predict the success rate for any $\vnr$ value of interest,
assuming linear correlation between $\rIi{\vvi}$ (for any $\vvi\in\sVinlier$) and $\vnr$,
and between $\rM$ and $\vnr$,
while assuming constant $|\sV|$.

\subsubsection{Baseline Approaches to Parameter Evaluation}
We determine the \emph{global success rate} through exhaustive evaluation of parameter values for the task of localizing the $500$ query images on the map.
The global success rate is an accurate representation of the actual localization capabilities of a method
and provides a good basis for parametrization decisions.
Any alternative approach to assessing the localization performance,
evaluated on the separate test images,
should enable us to make similar judgments about suitable parameter choices,
i.e. a similar trend between parameter values and localization performance should emerge.
Besides our prediction model,
we evaluate two other approaches to this task.

\begin{enumerate}
	\item \emph{Local success rate:}
	based on the previously introduced inlier and outlier evaluation,
	we propose a simpler model,
	which only compares voting peaks from the inlier evaluation to that of the outlier evaluation.
	In order to determine the local success rate,
	we evaluate how often it is the case that one of the voting peaks from the inlier evaluation receives at least two 
	inliers and overall more votes than any voting peak observed during the outlier evaluation.
	\item \emph{Inlier ratio:}
	the ratio of inliers among the matches might correlate with the success rate.
	Therefore, we use it as second alternative performance prediction.
	Again, this is computed with the inlier and outlier evaluation results.
\end{enumerate}
To predict localization performance with these two alternative approaches,
we perform inlier and outlier evaluation for each considered value of $\vnr$.

\subsubsection{Results}
\label{sec:results}
\begin{figure}[t]
	\vspace*{0.125cm}
	\begin{subfigure}[c]{0.328\columnwidth}
		\includegraphics[width=1\columnwidth]{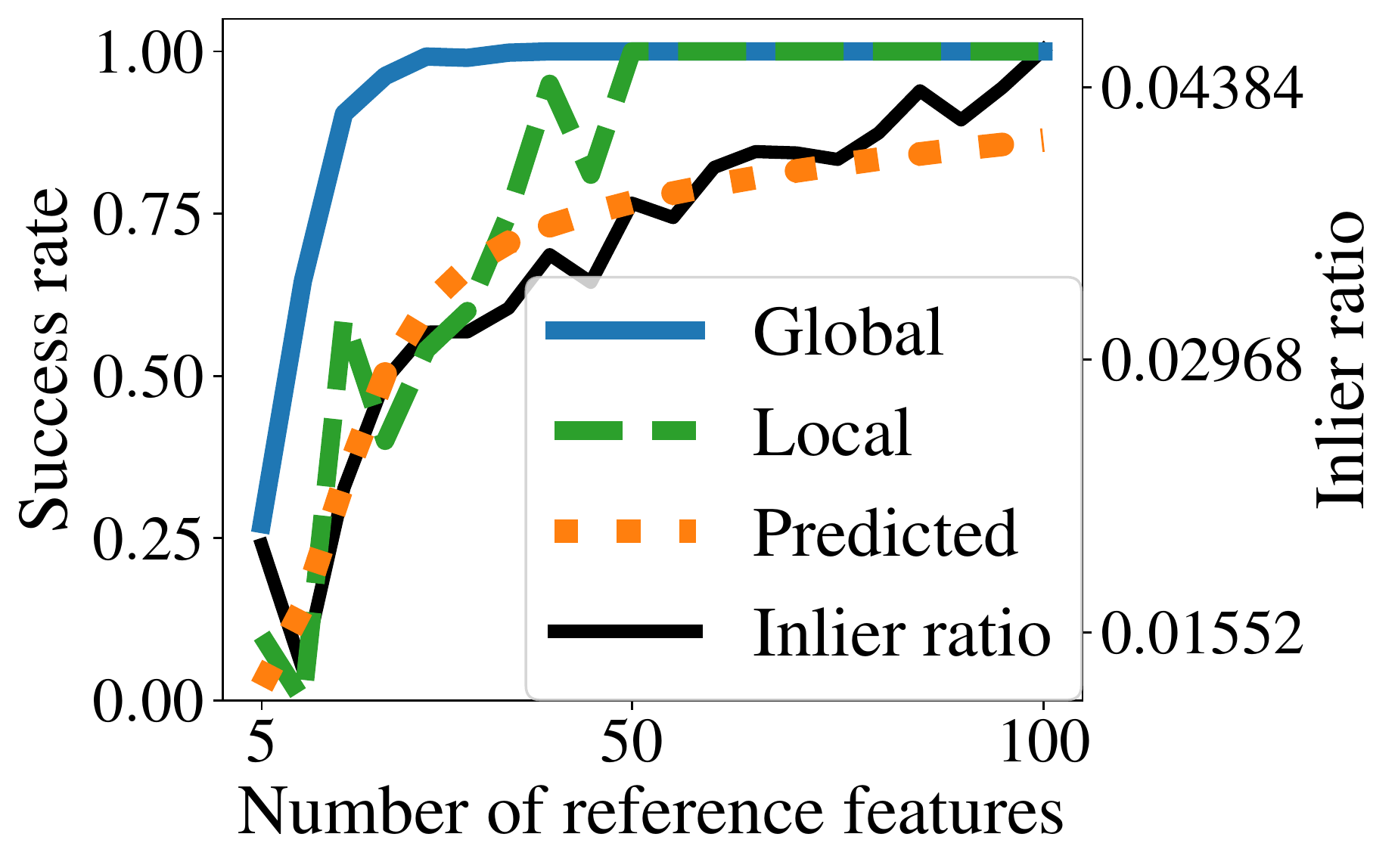}
		\subcaption{Tiles}
	\end{subfigure}
	\begin{subfigure}[c]{0.328\columnwidth}
		\includegraphics[width=1\columnwidth]{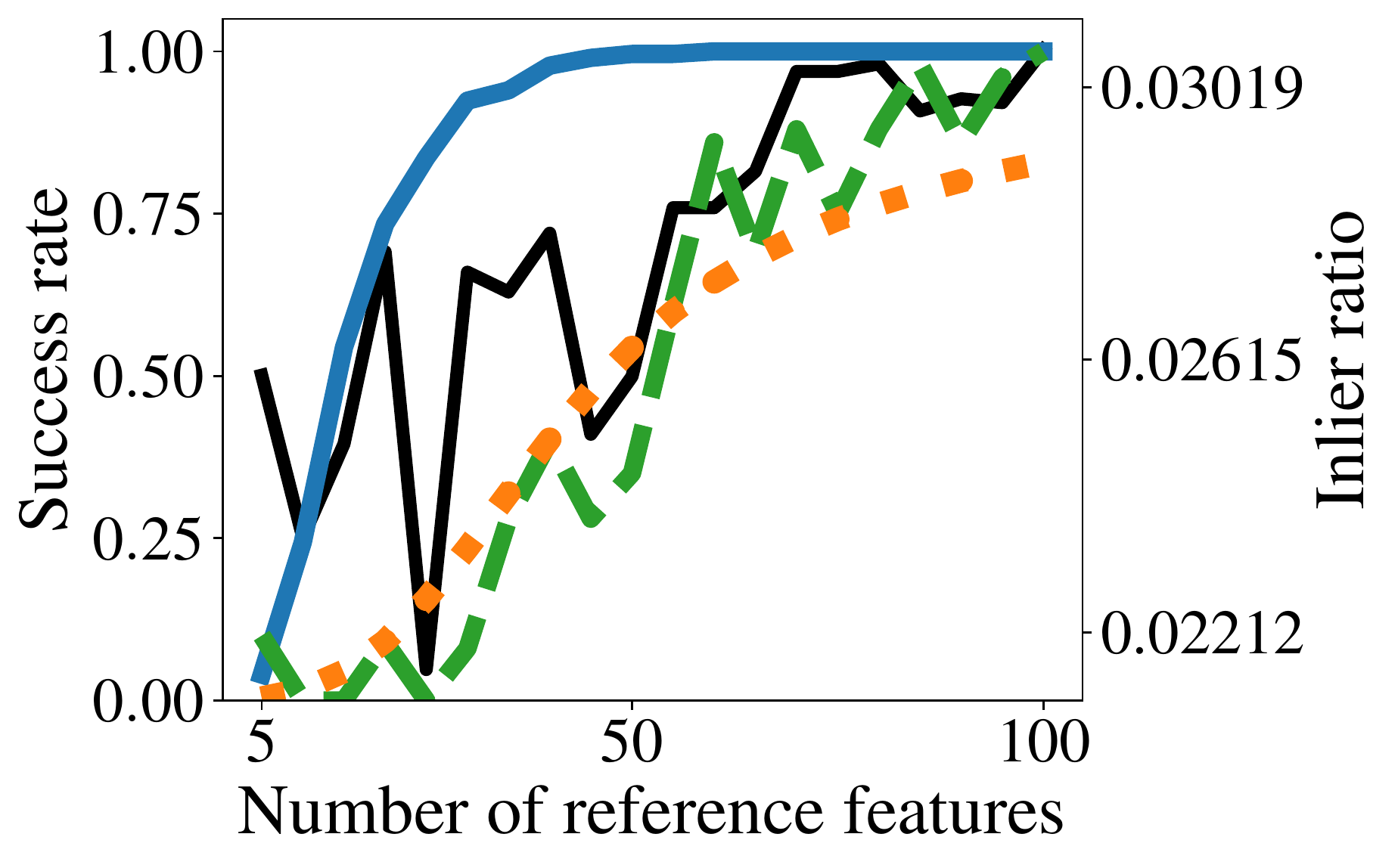}
		\subcaption{Asphalt (fine)}
	\end{subfigure}
	%\hfill
	\begin{subfigure}[c]{0.328\columnwidth}
		\includegraphics[width=1\columnwidth]{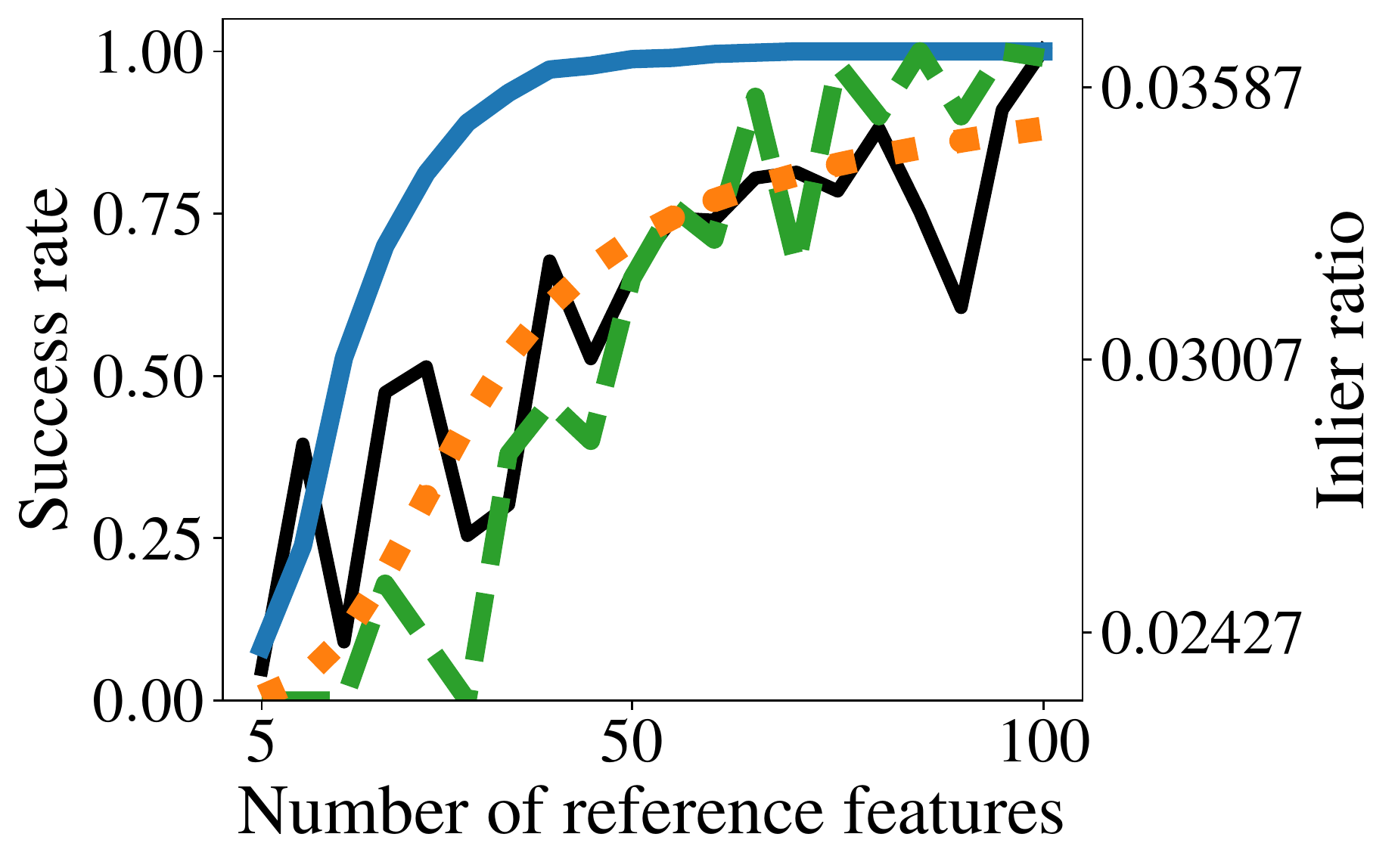}
		\subcaption{Asphalt (coarse)}
	\end{subfigure}
	\begin{subfigure}[c]{0.328\columnwidth}
		\includegraphics[width=1\columnwidth]{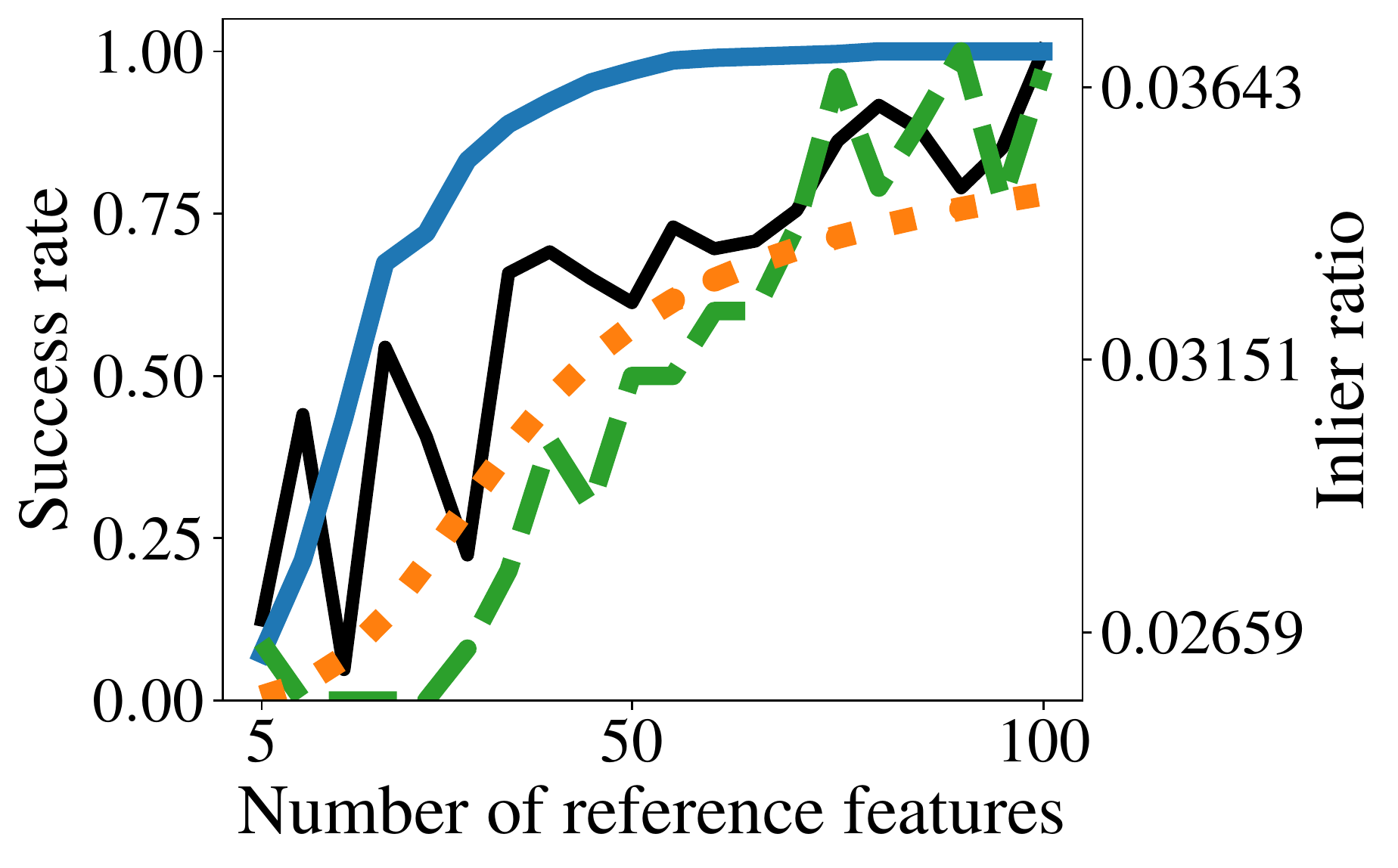}
		\subcaption{Carpet}
	\end{subfigure}
	%\hfill
	\begin{subfigure}[c]{0.328\columnwidth}
		\includegraphics[width=1\columnwidth]{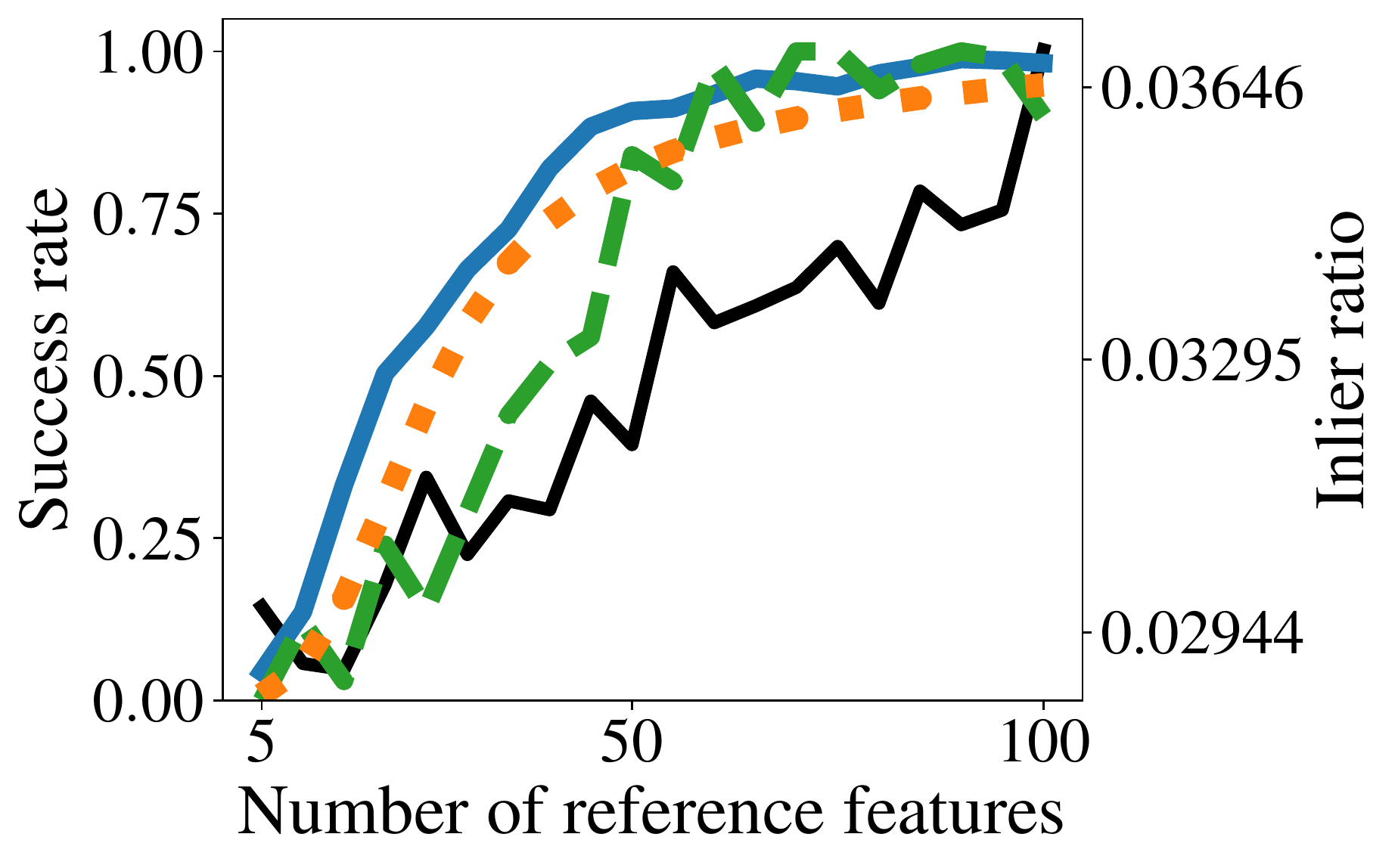}
		\subcaption{Concrete}
	\end{subfigure}
	%\hfill
	\begin{subfigure}[c]{0.328\columnwidth}
		\includegraphics[width=1\columnwidth]{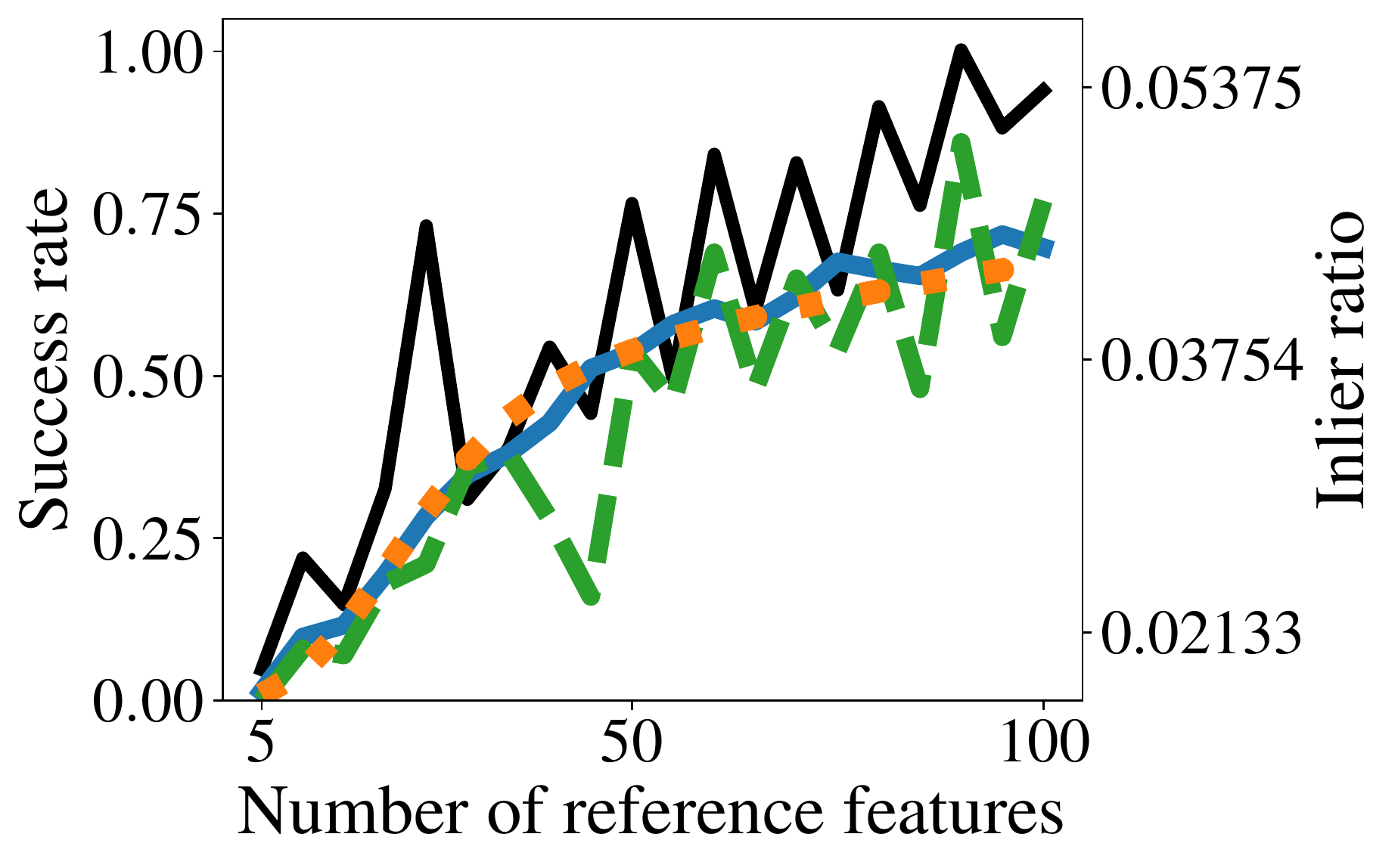}
		\subcaption{Wood}
	\end{subfigure}
	\caption{Global, local, and predicted success rates,
		as well as the inlier ratio, for Micro-GPS~\cite{Zhang_High-Prec-Localization} with varying $\vnr$ values.}
	\label{fig:results_mgps}	
\end{figure}

We evaluate Micro-GPS for $\vnr$ values ranging from $5$ to $100$ with increments of $5$,
and compute the average prediction errors over these $20$ evaluations.
For every texture type, we repeat this for $15$ different sets of test images,
each with $10$ query images and their overlapping reference images.
Overall, the prediction error of our model for the global success rate is on average $0.217$,
while it is $0.229$ for the local success rate.
Fig.~\ref{fig:results_mgps} presents texture-specific results, using one test image set respectively.
It also presents the inlier ratio.
Ideally, 
the curve of a performance indicator, 
should be similar to that of the global success rate.
However,
it is sufficient if it presents similar trends.
For example,
if the inlier ratio curve would present similar trends as that of the global success rate,
it would be suitable for parametrization.
But, we observe that,
while the general trend of the inlier ratio is often similar,
its curve is highly volatile,
e.g. for fine asphalt in a parameter value range between $5$ and $50$,
which could lead to suboptimal parametrization
or to a situation in which we get stuck in a local maximum.
The local success rate curves are more reliable.
But, only the curves of our model are smooth, monotonically increasing,
and present similar trends as the global success rate.
The predicted success rates saturate for larger values of $\vnr$ as for the global success rate,
which could lead to conservative parametrization choices.

\begin{figure}[t]
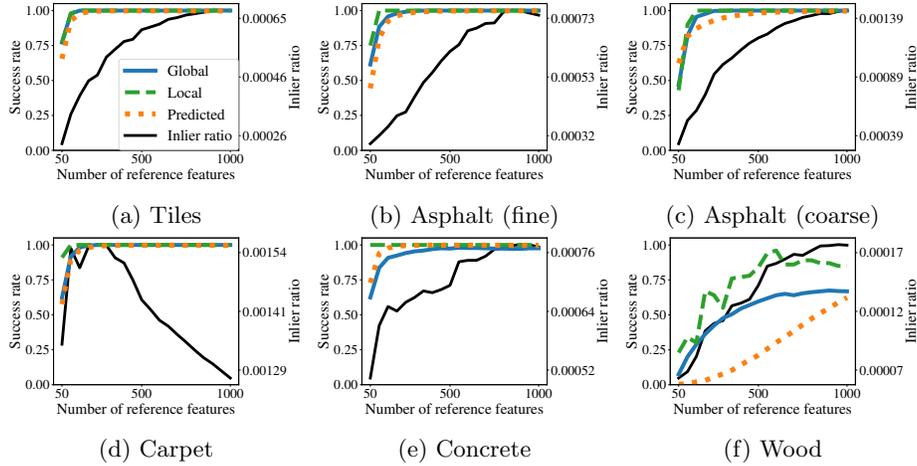

	\vspace*{0.125cm}
	\begin{subfigure}[c]{0.328\columnwidth}
		\includegraphics[width=1\columnwidth]{src/GCPR2020_IdentityMatchingApproach_nof_correct_matches_gt_cell_one_of_three/%
			success_rate_max_ref_keypoints_pred_vs_actual_IdentityMatchingApproach_tile2_10.pdf}
		\subcaption{Tiles}
	\end{subfigure}
	\begin{subfigure}[c]{0.328\columnwidth}
		\includegraphics[width=1\columnwidth]{src/GCPR2020_IdentityMatchingApproach_nof_correct_matches_gt_cell_one_of_three/%
			success_rate_max_ref_keypoints_pred_vs_actual_IdentityMatchingApproach_asphalt_10.pdf}
		\subcaption{Asphalt (fine)}
	\end{subfigure}
	%\hfill
	\begin{subfigure}[c]{0.328\columnwidth}
		\includegraphics[width=1\columnwidth]{src/GCPR2020_IdentityMatchingApproach_nof_correct_matches_gt_cell_one_of_three/%
			success_rate_max_ref_keypoints_pred_vs_actual_IdentityMatchingApproach_coarse_10.pdf}
		\subcaption{Asphalt (coarse)}
	\end{subfigure}
	\begin{subfigure}[c]{0.328\columnwidth}
		\includegraphics[width=1\columnwidth]{src/GCPR2020_IdentityMatchingApproach_nof_correct_matches_gt_cell_one_of_three/%
			success_rate_max_ref_keypoints_pred_vs_actual_IdentityMatchingApproach_carpet_10.pdf}
		\subcaption{Carpet}
	\end{subfigure}
	%\hfill
	\begin{subfigure}[c]{0.328\columnwidth}
		\includegraphics[width=1\columnwidth]{src/GCPR2020_IdentityMatchingApproach_nof_correct_matches_gt_cell_one_of_three/%
			success_rate_max_ref_keypoints_pred_vs_actual_IdentityMatchingApproach_concrete_10.pdf}
		\subcaption{Concrete}
	\end{subfigure}	
	%\hfill
	\begin{subfigure}[c]{0.328\columnwidth}
		\includegraphics[width=1\columnwidth]{src/GCPR2020_IdentityMatchingApproach_nof_correct_matches_gt_cell_one_of_three/%
			success_rate_max_ref_keypoints_pred_vs_actual_IdentityMatchingApproach_wood_10.pdf}
		\subcaption{Wood}
	\end{subfigure}
	\caption{Global, local, and predicted success rates,
		and inlier ratio, for Schmid et al.'s method~\cite{Schmid_GTBL} with 
		varying $\vnr$ values.}
	\label{fig:results_ours}	
\end{figure}

We evaluate Schmid et al.'s localization method for $\vnr$ values ranging from $50$ to $1000$ with increments of $50$.
Again, evaluation is repeated for $15$ test image sets.
The average error of our model is $0.058$, and $0.049$ for the local success rate.
Fig.~\ref{fig:results_ours} presents some of the results.
Our model is accurate for all textures, but wood (Fig.~\ref{fig:results_ours}(f)).
Closer analysis shows that this is caused by an underestimation of the globally observed number of inliers.
So, in this case, the test images were not sufficiently representative for the application area.
Apart from wood, the curves of the local success rate are similar to that of the global success rate.
However, for concrete, fine asphalt, and carpet,
the local success rate overestimates the performance of small $\vnr$ values.
The inlier ratio curves tend to saturate only for significantly larger $\vnr$ values as for the global success rate.

So far,
to evaluate our model,
we estimated expected global numbers of inliers and outliers based on a single parameter evaluation of $\vnr$
($\vnr=50$ for Micro-GPS and $\vnr=850$ for Schmid et al.'s method),
and based on the assumption of linear correlation between $\vnr$ and the number of inliers and outliers.
If, instead, we evaluate the inlier and outlier evaluation for every considered value of $\vnr$,
as it is done for the local success rate and the inlier ratio,
our average prediction error decreases slightly to $0.199$ for Micro-GPS
and to $0.048$ for Schmid et al.'s method.
However, the prediction curves are no longer monotonically increasing.

\subsection{Using the Model for Parameter Optimization}
\label{sec:evaluation:optimization_framework}
We propose a simple parameter optimization framework,
which uses our success rate prediction model to evaluate possible parameter settings,
and apply it to find texture-dependent parameter settings for Schmid et al.'s method~\cite{Schmid_GTBL}.

We select ten important  
parameters to be optimized:
$|\sFq|$, $\vnr$, the histogram cell size of the voting procedure,
the number of considered LATCH bits,
four parameters of the SIFT keypoint detector,
and two parameters of the LATCH descriptor.
For every parameter, a value range and a step size is defined.
This results in a parameter space with more than $5$ billion possible configurations.

The optimization framework continuously samples random parameter settings from the parameter space,
and evaluates them using the prediction model. % as previously described.
This means that the localization method is parametrized according to the selected parameter setting,
before inlier and outlier evaluations are performed on $10$ consecutive query images and their respective reference images
to obtain empiric observations,
which are then used to estimate the global success rate.
Whenever the predicted success rate of a sampled parameter setting is not lower than the predicted success 
rate of the previously best parameter set minus $0.05$,
a local optimization is performed for that setting.
Here, one of the ten parameters is randomly selected to be optimized,
and four different values (two larger and two smaller ones) are tested for it.
The four generated parameter settings are then evaluated on the test images,
except when the number of extracted reference image features $\vnr$ was chosen to be optimized.
In that case,
we employ the previously described approach of predicting the impact of the parameter change on the success rate.
This local optimization procedure is repeated as long as it increases the predicted method performance,
but at least $12$ times.

Our optimization framework keeps track of the best-performing parameter setting,
considering the predicted success rate and the computation time:
a setting is considered superior to another one if its predicted success rate
is at least $0.005$ higher,
or if it is not more than $0.005$ smaller but is faster to compute.

\subsubsection{Results}
We run the optimization separately on all six ground textures types for $12$ hours on a E3-1270 Intel Xeon 
CPU at $3.8$\,GHz.
For the final best-performing parameter configurations,
the global success rate is evaluated and
compared with the results using the default configuration suggested by the authors~\cite{Schmid_GTBL}.
The optimized configurations are very competitive with the default configuration:
the mean success rate over all six textures is $94.0\%$ compared to $94.3\%$ with the default,
while the mean localization time is reduced from $0.997$\,s with the default to $0.752$\,s.
We also evaluate $225$ randomly sampled configurations per texture,
observing a success rate of $87.1\%$ with a mean localization time of $0.771$\,s.
Particularly for wood,
guided parametrization is crucial,
as the mean success rate of randomly sampled configurations is $29.7\%$ compared to $68.6\%$ with our optimized 
configuration and $68.2\%$ with the default.
Some patterns can be observed in the optimized configurations.
On the more difficult textures wood and concrete,
the optimized configurations decreased the number of considered LATCH-bits from $15$ to $14$,
while this number is increased to $16$ for the other textures.
For wood, the most difficult texture,
the number of extracted reference and query features are increased from $850$ to $1000$,
respectively from $850$ to $950$,
while on the other textures these numbers are decreased, 
e.g. to $500$, respectively $600$, on fine asphalt,
reducing the computation time and memory consumption.

On average it took $20.1$\,s to evaluate a parameter configuration during the optimization procedure.
Further speedup is possible through parallelization.
In comparison,
evaluating the final configuration with all reference images to obtain the global success rate took on average 
$3347.6$\,s.
Our prediction model thus allows an acceleration by a factor of about $166$ in the evaluation of configurations.

\subsubsection{Discussion}
It is not to be expected that the automatically found parameters are better than those that have been determined in a laborious week-long manual process,
as in the case of Schmid et al.~\cite{Schmid_GTBL}.
The parameter space studied is the same,
so both approaches can find good solutions.
The advantage of the automated approach should be that a good configuration is found in shorter time and with less effort.
This goal is already achieved by our simple parameter optimization framework,
due to the employment of the prediction model.

\section{Conclusion}
\label{sec:conclusion}
We proposed a success rate prediction model for ground texture based localization methods,
and used it for a parameter optimization framework.
On the example of the number of extracted features per reference image $\vnr$,
we have shown that the predictions are sufficiently accurate for parametrization.
Furthermore, due to our model-based approach,
it is not necessary to fully evaluate every considered parameter configuration,
because for some parameters such as $\vnr$,
we can accurately estimate its impact on the model input values.

Our prediction model can be used to optimize any localization method parameter influencing the localization performance.
Accordingly, we were able to build a parameter optimization framework with it,
which can quickly evaluate any considered parameter configuration.
Using the configurations obtained from the framework,
we achieved a similar localization success rate as with the original default parameter setting,
while the localization time was significantly reduced.

%%%%%%%%%%%%%%%%%%%%%%%%%%%%%%%%%%%%%%%%%%%%%%%%%%%%%%%%%%%%%%%%%%%%%%%%%%%%%%%%

\bibliographystyle{splncs04}
\bibliography{./bib}

\begin{thebibliography}{10}
\providecommand{\url}[1]{\texttt{#1}}
\providecommand{\urlprefix}{URL }
\providecommand{\doi}[1]{https://doi.org/#1}

\bibitem{Avrithis_SpatialVerification}
Avrithis, Y., Tolias, G.: Hough pyramid matching: Speeded-up geometry
  re-ranking for large scale image retrieval. International Journal of Computer
  Vision (IJCV)  \textbf{107}(1),  1--19 (2014)

\bibitem{opencv_library}
Bradski, G.: The {OpenCV} library. Dr. Dobb's Journal of Software Tools  (2000)

\bibitem{chenstreetmap}
{Chen}, X., {Vempati}, A.S., {Beardsley}, P.: {StreetMap} - mapping and
  localization on ground planes using a downward facing camera. In: IEEE/RSJ
  International Conference on Intelligent Robots and Systems (IROS). pp.
  1672--1679 (Oct 2018)

\bibitem{Fang_intelligent-vehicles2}
Fang, H., Yang, M., Yang, R., Wang, C.: Ground-texture-based localization for
  intelligent vehicles. IEEE Transactions on Intelligent Transportation Systems
  (ITS)  \textbf{10}(3),  463--468 (Sept 2009)

\bibitem{Kelly_AGV}
Kelly, A., Nagy, B., Stager, D., Unnikrishnan, R.: Field and service
  applications - an infrastructure-free automated guided vehicle based on
  computer vision - an effort to make an industrial robot vehicle that can
  operate without supporting infrastructure. IEEE Robotics and Automation
  Magazine (RAM)  \textbf{14}(3),  24--34 (Sept 2007)

\bibitem{Kozak_Ranger}
Kozak, K.C., Alban, M.: Ranger: A ground-facing camera-based localization
  system for ground vehicles. In: IEEE/ION Position, Location and Navigation
  Symposium (PLANS). pp. 170--178 (April 2016)

\bibitem{Levi_LATCH}
Levi, G., Hassner, T.: {LATCH}: Learned arrangements of three patch codes. In:
  IEEE Winter Conference on Applications of Computer Vision (WACV). pp.~1--9
  (2016)

\bibitem{Lowe_SIFT2}
Lowe, D.G.: Distinctive image features from scale-invariant keypoints.
  International Journal of Computer Vision (IJCV)  \textbf{60}(2),  91--110
  (Nov 2004)

\bibitem{Mount_CoverageSelection}
Mount, J., Dawes, L., Milford, M.: Automatic coverage selection for
  surface-based visual localization. In: IEEE/RSJ International Conference on
  Intelligent Robots and Systems (IROS) (Nov 2019)

\bibitem{Muja_FLANN}
Muja, M., Lowe, D.G.: Fast approximate nearest neighbors with automatic
  algorithm configuration. In: {International Conference on Computer Vision
  Theory and Application (VISSAPP)}. pp. 331--340. INSTICC Press (2009)

\bibitem{Nagai_Path_Tracking}
Nagai, I., Watanabe, K.: Path tracking by a mobile robot equipped with only a
  downward facing camera. In: IEEE/RSJ International Conference on Intelligent
  Robots and Systems (IROS). pp. 6053--6058 (Sept 2015)

\bibitem{Schmid_Survey}
Schmid, J.F., Simon, S.F., Mester, R.: Features for ground texture based
  localization - a survey. In: Proceedings of the British Machine Vision
  Conference (BMVC) (2019)

\bibitem{Schmid_GTBL}
Schmid, J.F., Simon, S.F., Mester, R.: Ground texture based localization using
  compact binary descriptors. In: IEEE International Conference on Robotics and
  Automation (ICRA). pp. 1315--1321 (2020)

\bibitem{Schonberger_SpatialVerification}
Sch{\"o}nberger, J.L., Price, T., Sattler, T., Frahm, J.M., Pollefeys, M.: A
  vote-and-verify strategy for fast spatial verification in image retrieval.
  In: Lai, S.H., Lepetit, V., Nishino, K., Sato, Y. (eds.) Asian Conference on
  Computer Vision (ACCV). pp. 321--337. Springer International Publishing, Cham
  (2017)

\bibitem{Zeisl_SpatialVerification}
Zeisl, B., Sattler, T., Pollefeys, M.: Camera pose voting for large-scale
  image-based localization. In: IEEE International Conference on Computer
  Vision (ICCV) (December 2015)

\bibitem{Zhang_High-Prec-Localization}
{Zhang}, L., {Finkelstein}, A., {Rusinkiewicz}, S.: High-precision localization
  using ground texture. In: IEEE International Conference on Robotics and
  Automation (ICRA). pp. 6381--6387 (2019)

\end{thebibliography}

\clearpage

\title{Supplementary Material}
\titlerunning{Model-Based Parameter Optimization for GTBL Methods}
\author{Jan Fabian Schmid\inst{1,2} \and
	Stephan F. Simon\inst{1} \and
	Rudolf Mester\inst{3,2}}
\authorrunning{J.F. Schmid et al.}	
\institute{Robert Bosch GmbH, Hildesheim, Germany\\
	\email{SchmidJanFabian@gmail.com} \and
	VSI Lab, CS Dept., Goethe University, Frankfurt am Main, Germany \and
	Norwegian Open AI Lab, CS Dept., NTNU Trondheim, Norway
}

\maketitle

\begin{abstract}
Fig.~\ref{fig:example_images} shows some example images of the employed ground texture image database of Zhang et al.~\cite{Zhang_High-Prec-Localization}.
Section~\ref{sec:model_sup} presents the complete derivation of our localization success rate prediction model.
Then,
Section~\ref{sec:evaluation:inliers} presents our approach to determine the number of inliers that occurred during a localization attempt.
\end{abstract}

\begin{figure}[tb]
	\vspace*{0.125cm}
	\centering
	\includegraphics[width=0.15613\columnwidth]{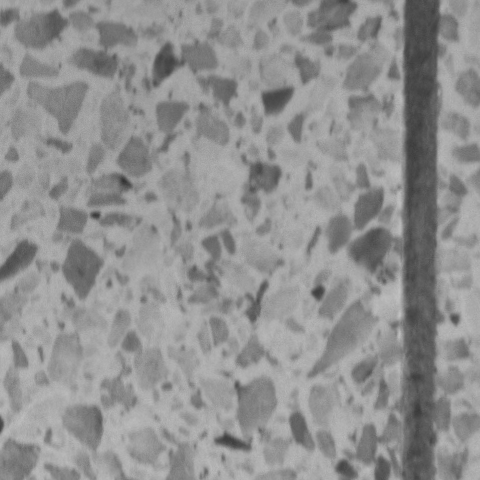}
	\includegraphics[width=0.15613\columnwidth]{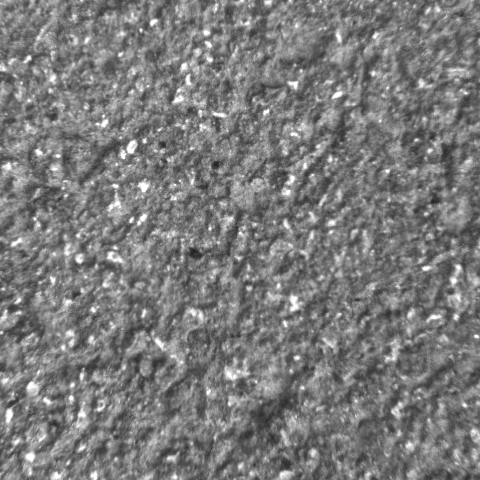}
	\includegraphics[width=0.15613\columnwidth]{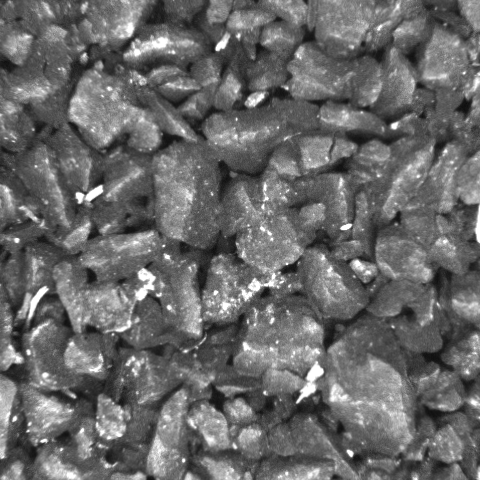}
	%%\vspace{0.1cm}
	\includegraphics[width=0.15613\columnwidth]{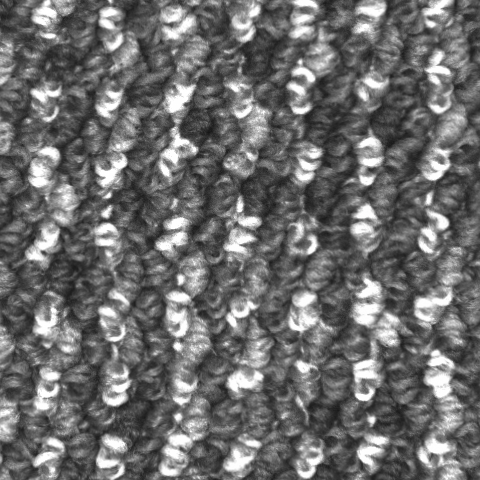}
	\includegraphics[width=0.15613\columnwidth]{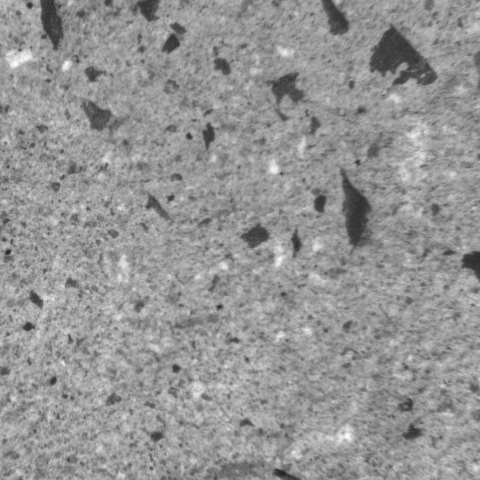}
	\includegraphics[width=0.15613\columnwidth]{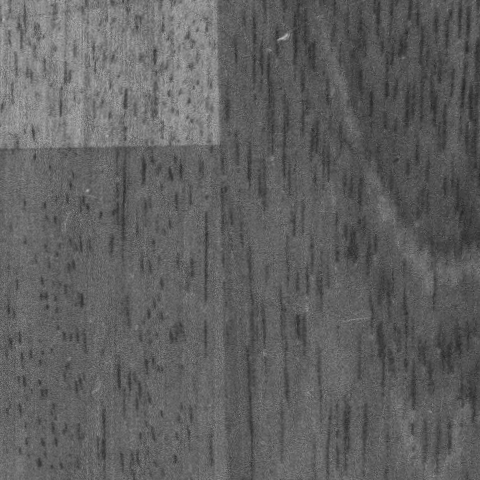}
	\caption{Images from the Zhang et al. database~\cite{Zhang_High-Prec-Localization}.
		Here, we show six different ground textures: tiles, fine asphalt, coarse 
		asphalt, carpet, concrete, and wood.}
	\label{fig:example_images}
\end{figure}

\section{Derivation of the Prediction Model}
\label{sec:model_sup}
We model the probability of successful localization for methods that have the properties described in 
Section $3$ of the main document.

Let the random variable $\rM$ denote the number of proposed feature matches, %from the feature matching step,
where $\rM=\rI+\rO$, with $\rI$ and $\rO$ denoting the numbers of inliers, respectively outliers.
Furthermore,
let $\sV$ denote the set of histogram voting cells that received at least one vote.
For a voting cell $\vv\in\sV$,
$\rMi{\vv}$ denotes the random variable that represents the number of votes cast onto it,
i.e. the number of matches with a corresponding pose estimate that projects the query image position into the boundaries
of voting cell $\vv$.
Similarly,
$\rIi{\vv}$ and $\rOi{\vv}$ represent respectively the number of inliers and outliers among them.
The voting peak is denoted as 
\begin{equation}
\vvp = \{\vv\in\sV|\rMi{\vv}=\max_{\vv'\in\sV}\rMi{\vv'}\}\,,
\end{equation}
assuming there is a single voting cell with most votes.

If we assume that the pose estimation algorithm works well,
localization succeeds when having two or more inliers on the voting peak,
as two correct matches are sufficient to determine the correct query image pose.
Generally, a single inlier is insufficient as it might be indistinguishable from an outlier.
Therefore, we model the localization success rate as
\begin{equation}
\Prob{\rIi{\vvp}\ge2}\,.
\end{equation}
Let $\sVinlier\subset\sV$ denote the subset of voting cells with at least one inlier vote.
If one of those voting cells receives two or more inlier votes and has overall more votes than any other voting cell,
localization succeeds.
For a voting cell $\vvi\in\sVinlier$,
we compute the probability of this condition being true as
\begin{equation}
\label{eq:prob_of_being_voting_peak_and_having_two_or_more_inliers}
\Prob{(\rIi{\vvi}\ge2)\cap(\rMi{\vvi}=\rMi{\vvp})}\,,
\end{equation}
where $\Prob{A\cap B}$ denotes the probability of both events $A$ and $B$ to occur together.
Localization succeeds if it is \emph{not} the case that this condition is \emph{not} true for any %voting cell 
$\vvi\in\sVinlier$:
\begin{equation}
\begin{split}
\label{eq:localization_success}
&\Prob{\rIi{\vvp}\ge2}=\\
&1-\left(\prod_{\vvi\in\sVinlier}\left[1 - 
\left(\Prob{(\rIi{\vvi}\ge2)\cap(\rMi{\vvi}=\rMi{\vvp})}\right)\right]\right)\,.
\end{split}
\end{equation}
In order to compute the probability of having $\vj$ votes on voting cell $\vv\in\sV$,
we consider the number of inliers $\rIi{\vv}$ and  the number of outliers $\rOi{\vv}$ on it:
\begin{equation}
\label{eq:nof_matches_is_inliers_plus_outliers}
\Prob{\rMi{\vv}=\vj} = \sum_{\vk=0}^{\vj}\left[\Prob{\rIi{\vv}=\vk} \cdot 
\ProbGiven{\rOi{\vv}=\vj-\vk}{\rIi{\vv}=\vk}\right]\,,
\end{equation}
with $\ProbGiven{A}{B}$ denoting the conditional probability of $A$ given $B$.
As a next step, we make an assumption about the distribution of outlier votes.
Here, we assume to have \emph{complete spatial randomness} (CSR) among the voting positions,
i.e. the probability $\poutvote$ of any outlier match $\vm\in\sO$,
casting a vote on the voting cell $\vv$ is the same for any voting cell $\vv\in\sV$.
Furthermore, we assume to have many matches.
So, we can assume statistical independence for any two voting cells $\vv_1,\vv_2\in\sV$
\begin{equation}
\label{eq:votes_on_two_cells_independent}
\ProbGiven{\rMi{\vv_1}=\vi}{\rMi{\vv_2}=\vj} = \Prob{\rMi{\vv_1}=\vi}\,.
\end{equation}
Also, assuming to have significantly more outliers than inliers,
we approximate
\begin{equation}
\label{eq:nof_outliers_independent_of_nof_inliers}
\ProbGiven{\rOi{\vv}=\vi}{\rIi{\vv}=\vj} = \Prob{\rOi{\vv}=\vi}\,.
\end{equation}
Based on the CSR assumption,
$\rOi{\vv}$ for $\vv\in\sV$ is binomially distributed.
The distribution is characterized by the number of outliers $\rO$,
and the probability $\poutvote=1/|\sV|$ that each of them has for casting a vote on $\vv$:
\begin{equation}
\label{eq:prob_outvote}
\Prob{\rOi{\vv}=\vi}=\fB(\vi|\poutvote, \rO)\,,
\end{equation}
where $\fB(\vi|\mathrm{p}, \vn)$ denotes the probability of observing $\vi$ successes in $\vn$ independent Bernoulli 
trials,
each with a success probability of $\mathrm{p}$.

To estimate $\rIi{\vv}$,
the number of inliers casting a vote on $\vv\in\sV$,
we assume that every extracted query feature $\vfq\in\sFq$ has the same probability $\pinvotev{\vv}$ of 
generating one inlier vote on $\vv$
(with $\pinvotev{\vv}=0$ for any $\vv\in\sV\setminus\sVinlier$),
and we assume that no query feature will generate more than one inlier vote for $\vv$.
Accordingly, the random variable $\rIi{\vv}$ is binomially distributed as well,
depending on $\pinvotev{\vv}$ and the number of extracted query image features $|\sFq|$:
\begin{equation}
\label{eq:prob_invote}
\Prob{\rIi{\vv}=\vi}=\fB(\vi|\pinvotev{\vv}, |\sFq|)\,.
\end{equation}

The upper limit for the number of votes any voting cell can receive is $\rM$.
Considering this and Eq.~(\ref{eq:votes_on_two_cells_independent}),
we estimate
\begin{equation}
\Prob{\rMi{\vvi}=\rMi{\vvp}}=
\sum_{\vj=1}^{\rM}\left[\Prob{\rMi{\vvi}=\vj}\cdot\!\!\!\!\!\prod_{\vv\in\sV\setminus\{\vvi\}}\!\!\!\!\!\Prob{\rMi{\vv}<\vj}\right]\,.
\end{equation}
Using Eq.~(\ref{eq:nof_matches_is_inliers_plus_outliers}) and %the approximation of 
Eq.~(\ref{eq:nof_outliers_independent_of_nof_inliers}),
we obtain
\begin{equation}
\begin{split}
&\Prob{\rMi{\vvi}=\rMi{\vvp}} = \\
&\sum_{\vj=1}^{\rM}\!\left[\sum_{\vk=0}^{\vj}\left[\Prob{\rIi{\vvi}=\vk}\cdot\Prob{\rOi{\vvi}=\vj-\vk}\right]\cdot
\!\!\!\!\!\!\prod_{\vv\in\sV\setminus\{\vvi\}}\!\!\!\!\!\Prob{\rMi{\vv}<\vj}\right].
%&\sum_{\vj=1}^{\rM}\left[\sum_{\vk=0}^{\vj}\left[\Prob{\rIi{\vvi}=\vk}\cdot\Prob{\rOi{\vvi}=\vj-\vk}\right]\cdot\right.\\
%&\left.\prod_{\vv\in\sV\setminus\{\vvi\}}\Prob{\rMi{\vv}<\vj}\right]\,.
\end{split}
\end{equation}
Now, we can use this in Eq.~(\ref{eq:prob_of_being_voting_peak_and_having_two_or_more_inliers}) to obtain
\begin{equation}
\label{eq:success_with_vi}
\begin{split}
&\Prob{(\rIi{\vvi}\ge2)\cap(\rMi{\vvi}=\rMi{\vvp})} = \\
&\sum_{\vj=2}^{\rM}\!\left[\sum_{\vk=2}^{\vj}\left[\Prob{\rIi{\vvi}=\vk}\cdot\Prob{\rOi{\vvi}=\vj-\vk}\right]\cdot
\!\!\!\!\!\!\prod_{\vv\in\sV\setminus\{\vvi\}}\!\!\!\!\!\Prob{\rMi{\vv}<\vj}\right].
\end{split}
\end{equation}
%Using Eq.~(\ref{eq:prob_outvote}) and Eq.~(\ref{eq:prob_invote}), we get
Finally, considering Eq.~(\ref{eq:success_with_vi}) with the substitution
\begin{equation}
\begin{split}
\Prob{\rMi{\vv}<\vj} &= \sum_{\vk=0}^{\vj-1}\Prob{\rMi{\vv}=\vk} \\
&=\sum_{\vk=0}^{\vj-1}\left[\sum_{\vl=0}^{\vk}\left[\Prob{\rIi{\vv}=\vl}\cdot\Prob{\rOi{\vv}=\vk-\vl}\right]\right]\,,
\end{split}
\end{equation}
we model the probability of observing a successful localization attempt using Eq.~(\ref{eq:localization_success})
as: 
\begin{equation}
\label{eq:model_sup}
\begin{split}
&\Prob{\rIi{\vvp}\ge2} = 1 - \vx,\\
\end{split}
\end{equation}
with
\begin{equation}
\begin{split}
\vx&=\prod_{\vvi\in\sVinlier}\left[1 - 
\left(\Prob{(\rIi{\vvi}\ge2)\cap(\rMi{\vvi}=\rMi{\vvp})}\right)\right]\\
%&=1-\left(\prod_{\vvi\in\sVinlier}\left[1 - 
%\left(\sum_{\vj=2}^{\rM}\left[\sum_{\vk=2}^{\vj}\left[\fB(\vk|\pinvotev{\vvi}, |\sFq|)\cdot\fB(\vj-\vk|\poutvote, 
%\rO)\right]\cdot \right.\right.\right.\right.\\
%&\left.\left.\left.\left. 
%\prod_{\vv\in\sV\setminus\{\vvi\}}\sum_{\vk=0}^{\vj-1}\left[\sum_{\vl=0}^{\vk}\left[\fB(\vl|\pinvotev{\vv}, 
%|\sFq|)\cdot\fB(\vk-\vl|\poutvote, \rO)\right]\right]\right]\right)\right]\right)\,.\\
&=\prod_{\vvi\in\sVinlier}\left[1 - 
\left(\sum_{\vj=2}^{\rM}\left[\sum_{\vk=2}^{\vj}\left[\Prob{\rIi{\vvi}=\vk}\cdot\Prob{\rOi{\vvi}=\vj-\vk}\right]\cdot 
\right.\right.\right.\\
&\left.\left.\left. 
\prod_{\vv\in\sV\setminus\{\vvi\}}\sum_{\vk=0}^{\vj-1}\left[\sum_{\vl=0}^{\vk}\left[\Prob{\rIi{\vv}=\vl}\cdot      
\Prob{\rOi{\vv}=\vk-\vl}\right]\right]\right]\right)\right]\,.
\end{split}
\end{equation}

\section{What is a Correct Match of Features?}
\label{sec:evaluation:inliers}
Counting inliers correctly is a key requirement for the use of the proposed prediction model.
We defined inliers as pairs,
consisting of query and reference feature, %$\vm\in\sM = (\vfq\in\sFq, \vfr\in \sFR)$,
that can be used for successful pose estimation.
To determine whether we are counting inliers correctly,
we observe the number of inliers on the voting peak of successful localization attempts.
Localization attempts without any inliers on the voting peak should not succeed.
One approach to determine the correctness of a match
would be to determine whether its corresponding pose estimate itself is already correct.
However, this underestimates the actual inlier count,
e.g. for Micro-GPS, on average there are less than $0.01$ matches per localization attempt satisfying this condition,
while it achieves a success rate of $90\%$.

Alternatively, we could take the employed pose estimation approach into account.
Both examined localization methods do not use the orientation information of keypoint objects for the final pose 
estimation.
Instead, they determine the query image pose using the voting positions of $2$ (or more) matches.
We propose three ways of counting inliers:
(1) the matches with correct corresponding pose estimate, not considering the orientation error;
(2) the matches that if paired with another match, voting for the same voting cell, create correct pose estimates;
(3) the matches that if paired with a fake keypoint object,
which is positioned on the ground truth query image position,
create correct pose estimates.
We observe similar inlier counts for all three proposed measures with averages of $7$ to $8$ inliers per localization 
attempt of Micro-GPS~\cite{Zhang_High-Prec-Localization}, respectively $76$ to $79$ for Schmid et al.'s method~\cite{Schmid_GTBL}.
Finally, we decide to treat any match as inlier for which any of these three measures has positive response.

\end{document}